%% file: main.tex
\documentclass{article}

\usepackage[nonatbib, final]{tackling_climate_workshop_style}
\usepackage[utf8]{inputenc} 
\usepackage[T1]{fontenc}    
\usepackage{hyperref}       
\usepackage{url}            
\usepackage{booktabs}       
\usepackage{amsfonts}       
\usepackage{amsmath}
\usepackage{nicefrac}       
\usepackage{wrapfig}
\usepackage{tabularray}
\usepackage{microtype}      
\usepackage{graphicx,float}

\usepackage[
backend=biber,
style=numeric,
sorting=ynt
]{biblatex}
\usepackage{graphicx}
\graphicspath{
    {img/}
}
\usepackage{subcaption}
\usepackage{rotating}
\usepackage[font=small,labelfont=bf]{caption}
\title{Domain Adaptation for Sustainable Soil Management using Causal and Contrastive Constraint Minimization }
\author{%
  {Somya Sharma}$^{a}$ \And
  {Swati Sharma}$^{b*}$  \And
  {Rafael Padilha}$^{b}$ \And 
  {Emre Kiciman}$^{b}$ \And {Ranveer Chandra}$^{b}$ \\ 
  \\ \texttt{swatisharma@microsoft.com}
  \\ a University of Minnesota Twin Cities, USA
  \\ b Microsoft Research, Redmond, USA
}

\begin{document}

\maketitle

\begin{abstract}
Monitoring organic matter is pivotal for maintaining soil health and can help inform sustainable soil management practices. While sensor-based soil information offers higher-fidelity and reliable insights into organic matter changes, sampling and measuring sensor data is cost-prohibitive. We propose a multi-modal, scalable framework that can estimate organic matter from remote sensing data, a more readily available data source while leveraging sparse soil information for improving generalization. Using the sensor data, we preserve underlying causal relations among sensor attributes and organic matter. Simultaneously we leverage inherent structure in the data and train the model to discriminate among domains using contrastive learning. This causal and contrastive constraint minimization ensures improved generalization and adaptation to other domains. We also shed light on the interpretability of the framework by identifying attributes that are important for improving generalization. Identifying these key soil attributes that affect organic matter will aid in efforts to standardize data collection efforts.

\end{abstract}




\section{Introduction}
\input{intro}
\section{Methods}

\input{method}

\section{Results}
\input{results}
\vspace{-0.2cm}

\section{Discussion and Future Work}
\input{conclusion}

\small
\printbibliography

\normalsize

\appendix
\include{appendix}

\end{document}

%% file: intro.tex
\noindent \paragraph{Background} Measuring and monitoring soil organic matter (OM) is pivotal in fighting against climate change \cite{droste2020soil}. Apart from contributing to improving soil health, organic matter has several co-benefits - such as fighting against soil erosion and preserving the water table ~\cite{SCHJONNING201835, soil-5-15-2019, Lal27A}. These characteristics make it imperative to preserve organic matter, especially in regions that are becoming vulnerable to droughts and landslides as the climate changes. While monitoring OM is an important problem, sampling and measuring OM (and other soil attributes) is expensive (e.g., the average cost of soil testing in the US is \$1444) ~\cite{Willis_2022}. Furthermore, to ensure accurate and reliable measurement of soil properties, it is necessary to conduct sampling at intervals of every tens of meters~\cite{kapetanovic2019farmbeats}.

\noindent \paragraph{Related Work} Alternatively, insights about OM can be inferred from other soil characteristics and remote sensing information using data-driven ML methods \cite{Heuvelink_2020, Nguyen_2021, Tang_2022, Zhang_2022, Sakhaee_2021, An_2022, Keskin_2019, Fathizad_2022, Meng_2022, Emadi_2020, Sakhaee_2022}. While sensor data input offers high-fidelity insights and captures the variability in the dominant soil processes in the region, collecting data from sensors is expensive. On the other hand, remote sensing data is inexpensive and widely available for different parts of the world. Notably, while remote sensing data are underutilized in low-income regions, the benefits derived from use of remote sensing will be the most in these regions where the effect of climate change is prominent\cite{rolf2021generalizable}. However, insights from remote sensing data may be biased due to noise and coarse-scale resolution \cite{boonprong2018classification, yanan2020cloud, zhao2023seeing}. In the context of OM mapping, domain adaptation has been useful in overcoming some of the challenges due to remote sensing data and in adapting models to differing land types and countries \cite{bursac2022instance, shen2022deep}. While these studies use local soil information as input variables, which may be costly to collect, the studies conclude that domain adaptation helps in improving the generalization to other domains or regions.

\noindent \paragraph{Summary of Contributions} We propose a scalable and generalizable domain adaptation framework for mitigating bias from remote sensing data using sparse sensor data as auxiliary data. Even if sensor data is not available during inference, leveraging sensor data while training can help capture the variability in the underlying soil processes and improve the generalization of the framework. Such a framework can be built on data-rich entities and be transferred to unobserved entities without further fine-tuning. Conventional ML methods overlook the underlying causal structure in the data, limiting their out-of-distribution generalization. To overcome this, we propose the use of causal constraint minimization to ensure that the relation between the sensor data attributes and OM is preserved in the posterior OM distribution across different regions. Traditional ML methods also struggle with generalization to locations with unobserved drivers (no training data). We leverage the spatial heterogeneity in the sensor attributes to improve generalization to out-of-distribution (OOD) locations by influencing embeddings via contrastive learning. Our analysis is also able to identify key soil attributes that affect organic matter, potentially improving understanding of how to optimize soil management practices and standardize data collection approaches. 


%% file: method.tex

The \textbf{backbone model} is a CNN autoencoder (Fig.~\ref{fig:backbone-model}) that estimates OM from satellite imagery data (more details in Appendix).  Although this model uses satellite imagery input, which is more readily available and captures information about changing soil properties, vegetation, and climate, leveraging soil attributes can help distill the important signals from the images and improve the encodings learned by the autoencoder framework. To achieve this, we propose incorporating two regularization schemes - causal constraint minimization and contrastive learning.

\begin{figure*}
    \centering
    \includegraphics[width=0.9\linewidth]{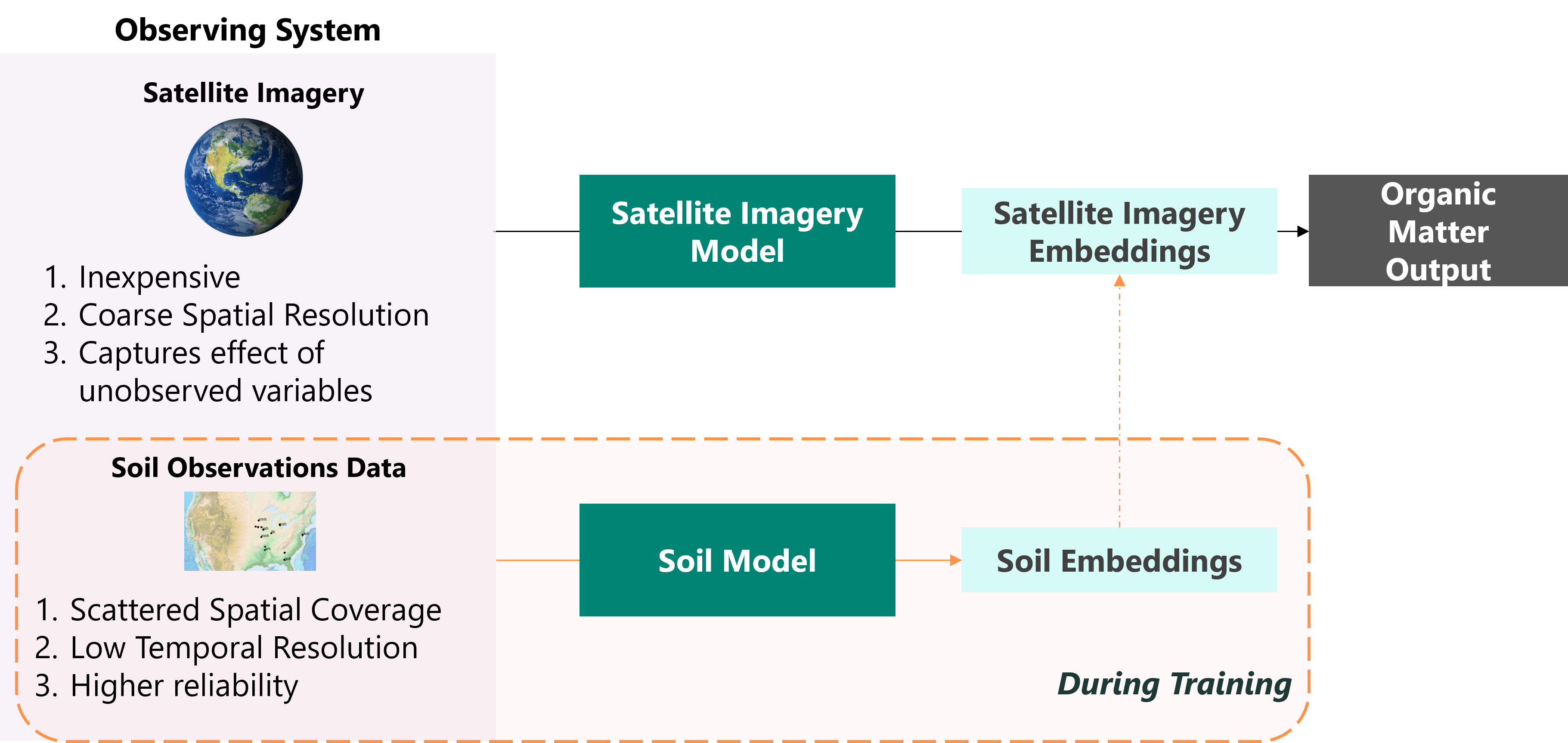}
    \caption{Domain Adaptation framework for Organic Matter Modeling using Causal and Contrastive Constraint Minimization. Conditioning the satellite image embedding using soil attribute embeddings provides additional context or guidance on how the underlying soil properties impact OM.}
    \label{fig:model-diagram}
\end{figure*}


Causally adaptive constraint minimization (CACM) has been implemented to improve generalization under distributional shifts in computer vision benchmark datasets \cite{kaur2022modeling}. The method utilizes different independence constraints based on how attributes in training data relate to response variables. These constraints are incorporated in the loss function to regularize the training by adaptively enforcing the correct independence constraints. We extend the \textbf{causal constraint minimization} framework to a setting with continuous attributes and continuous response variable on real-world dataset. We modify the framework to incorporate regularization loss terms that ensure that the encoding space follows the distributional properties that reflect the causal relations among sensor attributes and OM instead of the output from the autoencoder. This helps ensure that the output from the model is not being over-smoothed by the constraint minimization. Following the CACM framework, our model also incorporates three types of causal relations - confounded with OM, caused by OM, or independent of OM. For attributes that are independent of OM, we enforce $\sum_{i=1}^{|A_{ind}|} \sum_{j>i}MMD(P(\phi(x)|a_{i,ind}),P(\phi(x)|a_{j,ind}))$. For attributes that are caused by OM, we enforce $\sum_{i=1}^{|A_{cause}|} \sum_{j>i}MMD(P(\phi(x)|a_{i,cause},y),P(\phi(x)|a_{j,cause},y))$. For attributes that are confounded with OM via a confounding variable, we enforce,$\sum_{\|E\|}\sum_{i=1}^{|A_{conf}|}\sum_{j>i}MMD(P(\phi(x)|a_{i,conf}),P(\phi(x)|a_{j,conf}))$. Here, $\phi(x)$ are the encodings obtained from the satellite imagery CNN encoder. To compute the conditional probabilities, for each attribute $a$, we categorize the samples by attribute value below and above the mean attribute value. Here, $A_{ind}$, $A_{cause}$ and $A_{conf}$ refer to the sets containing independent, causal and confounded attributes, respectively. Adaptively choosing the regularization term depending on the causal relation enables us to ensure that the independence constraints are also reflected in the conditional embedding distribution.


We incorporate \textbf{contrastive learning} using another ANN autoencoder to learn representations for the soil attributes. The embeddings of this ANN model are used to provide additional context on how the underlying soil properties can impact the OM distribution. By incorporating contrastive learning, we capture location-specific patterns while learning to discriminate between the variations associated with different farming environments. We define positive and negative pairs based on differing locations. Samples from the same location are expected to have similar characteristics, including soil type, management practices, and environmental conditions, while different locations are more likely to have dissimilar characteristics. We include triplet loss or contrastive loss during training as $\mathcal{L}_{\text{Contrastive}} = D(z_a, z_p) - D(z_a, z_n)$. Here, $z_a$, $z_p$, and $z_n$ refer to the embeddings from the anchor domain, positive pair of the domain, and negative pair of the domain. Here, we use Euclidean distance as our distance metric $D$. In each iteration, for each location, we randomly sample positive and negative pairs to regularize the loss. Here, the positive pair refers to a sample from the same location collected in a different year, and the negative pair refers to a sample from a different location.

\begin{figure}[h]
    \centering
    \includegraphics[width=0.9\textwidth]{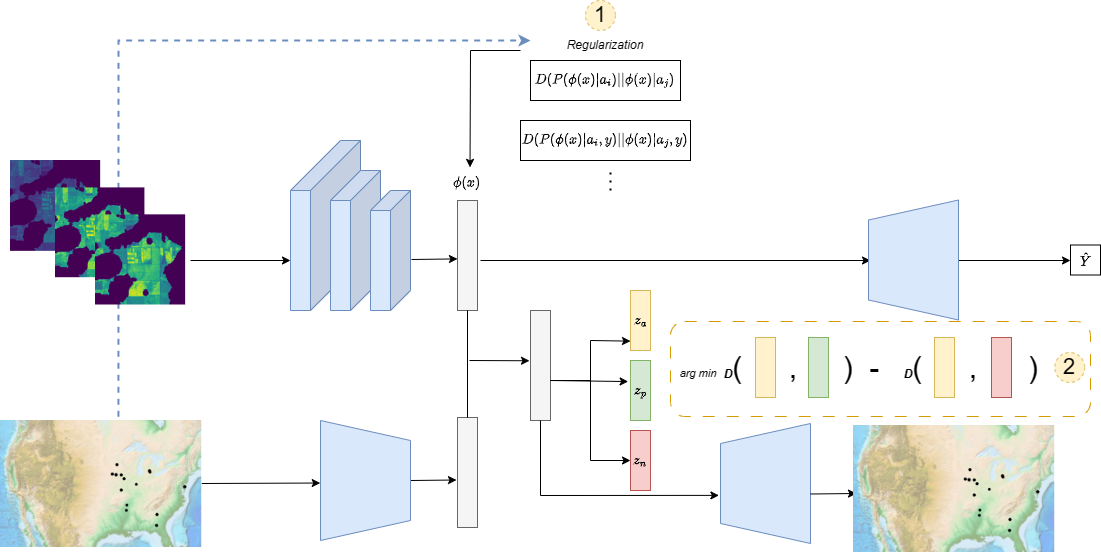}
    \caption{Domain Adaptation framework for Organic Matter Modeling using Causal and Contrastive Constraint Minimization. The bi-level optimization scheme first enforces causal independence constraints and then modifies the embeddings via contrastive learning.}
    \label{fig:model}
\end{figure}

The Causal and Contrastive Constraint Minimization approach is shown in Fig.~\ref{fig:model}.

%% file: results.tex
\noindent \paragraph{Data}
The satellite imagery data is obtained from \href{https://www.esa.int/Applications/Observing_the_Earth/Copernicus/Sentinel-2}{Sentinel-2}. The sensor data is obtained from publicly available \href{https://www.genomes2fields.org/home/}{Genomes to Fields (G2F) dataset}. The sensor data includes information about soil attributes (soil texture, micronutrients, fertilizer application), management practices and also the main variable of interest, OM. 
More details on included locations, datasets, and preprocessing are given in the Appendix.

\vspace{-0.4cm}

\noindent \paragraph{Generalization}


To test the OOD generalization of the framework, we train and test on different locations. Table~\ref{tab:ood} presents results on OOD generalization results where the test set includes eastern US locations ( Georgia and  Delaware) while the train set includes locations from the western US. The table compares models that use only satellite data as input with gold standard models that are also able to use sensor data. While using sensor data as input improves generalization, sensor data may be expensive to collect and use during model inference. Therefore, we propose leveraging sensor data to improve embeddings in our model using causal constraint minimization. In such a scenario, sensor data is only used while training the model as an auxiliary dataset that influences encodings.  We can further evaluate if we can leverage spatial heterogeneity among the different domains to improve generalization. We leverage contrastive learning to learn encodings that enable us to maximize the similarity between samples from the same location and maximize the dissimilarity between samples from different locations. The results suggest that apart from leveraging contrastive learning, successive implementation of causal and contrastive constraint minimization enables the model to improve OM estimation.

\vspace{-0.3cm}

\noindent \paragraph{Domain Adaptation}


\begin{table}
\centering
\scriptsize
\hspace{-1.8cm}
\subfloat[Out of Distribution Generalization]   
{%
\begin{tabular}{c|c|p{1.1cm}|c}
    \toprule
         Model & Input Data & Auxiliary Training Data & MSE   \\
     \midrule
         Random Forest & Satellite Data & - &   0.6351 \\
         Random Forest & Satellite, Sensor & - & 0.2180 \\
         CNN & Satellite & - & 0.3297 \\
         CNN & Satellite, Sensor & - & 0.0814 \\
         CNN$_{\text{CACM}}$ & Satellite & Sensor & 0.0578 \\
         CNN$_{\text{CACM + Contrastive}}$ & Satellite & Sensor & 0.0513 \\
         CNN$_{\text{Contrastive}}$ & Satellite & Sensor & 0.0657\\
         \bottomrule
    \end{tabular}
         \label{tab:ood}

    }
    \hspace{0.1cm}
\subfloat[Domain Adaptation using CNN backbone model]     
{%
    \begin{tabular}{c|c|c|c }
    \toprule
         Fine-Tuning Env.  & ${\text{CACM}}$ & ${\text{CACM+Contrastive}}$ & ${\text{Contrastive}}$   \\
         \midrule
         - &   0.3072 & 0.2354 & 0.1637 \\
         random & 0.1998 & 0.1702& 0.1153  \\
         closest & 0.0688 & 0.0509 & 0.0923\\
         farthest & 0.1134 & 0.0977 & 0.1150 \\
         \bottomrule
    \end{tabular}
\label{tab:finetune}
}
\caption{Results on OOD generalization and Domain Adaptation}

\end{table}

To further test how the out-of-distribution generalization can benefit from domain adaptation, we evaluate the model performance in K-fold cross-validation splits where K is the number of different states. The value of K is 6 in these experiments. Table~\ref{tab:finetune} reports how model performance is impacted by fine-tuning the model in different locations relative to the test environment. For each of the K states, we subset one state as fine-tuning location. This way (K-2) locations are used for pertaining to the model, one location is used to fine-tune the model, and one location is used to test in each of the folds. We report the average test MSE for each test split in Table~\ref{tab:fintune_closest}. The table shows that fine-tuning using constrain minimization is useful as opposed to not using any fine-tuning data. This may suggest that fine-tuning on another domain allows the model to escape any local minima that is achieved without fine-tuning. Among the three ways of choosing the fine-tuning location, when the closest location is used for fine-tuning, the framework is able to improve generalization. This may be because locations that are closely located may have similar physical and soil farm attributes, improving the transfer of knowledge between domains. As opposed to fine-tuning based on a randomly selected location, fine-tuning on the farthest location results in improved performance. It has been shown that training on heterogeneous locations improves OOD generalization \cite{willard2023time}. Training on the farthest locations enables the model to learn to discriminate among different locations. Since causal constraint minimization enables the framework to preserve the underlying relation among the dominant soil processes,  the improvement of farthest location-based fine-tuning over random location and no fine-tuning is more evident. 

\begin{table}[h]
    \centering
    \scriptsize
    \begin{tabular}{c|c|c|c|c|c|c|c|c} 
    \toprule
        Test Environment & & Delaware & Georgia & Germany & Iowa & Illinois & Michigan &\\
        Fine Tune Environment & & Georgia & Illinois & Delaware & Illinois & Iowa & Illinois & Average \\
        \midrule
      CNN$_{\text{CACM + Contrastive}}$ & -  & 0.1843& 0.1327 & 0.3265 &0.6043 & 0.0716 &0.0927 & 0.2354 \\
      CNN$_{\text{CACM + Contrastive}}$ & FT  & 0.0235&\bf{0.0336} & \bf{0.0131}&0.1528 &0.0418 &0.0404 & \bf{0.0509}\\
      CNN$_{\text{CACM}}$   & - & 0.0491& 0.0406& 0.3921& 1.2433&0.0796  & 0.0382 & 0.3072\\
      CNN$_{\text{CACM}}$   & FT & 0.0552&0.2127 &0.0136 &\bf{0.0715} &\bf{0.0309}  & 0.0289 & 0.0688\\
      CNN$_{\text{Contrastive}}$   & - & \bf{0.0195} & 0.0338 & 0.1049 &0.7750 &0.0380 &0.0112 & 0.1637 \\
      CNN$_{\text{Contrastive}}$   & FT & \bf{0.0195} & 0.0367 &0.0162 & 0.1534 & 0.3171 & \bf{0.0110} & 0.0923 \\
      \bottomrule
    \end{tabular}
    \caption{Domain Adaptation Model Performance on closest environment. FT: Fine-tuned}
    \label{tab:fintune_closest}
\end{table}

\begin{figure}
    \centering
    \includegraphics[width=0.7\textwidth]{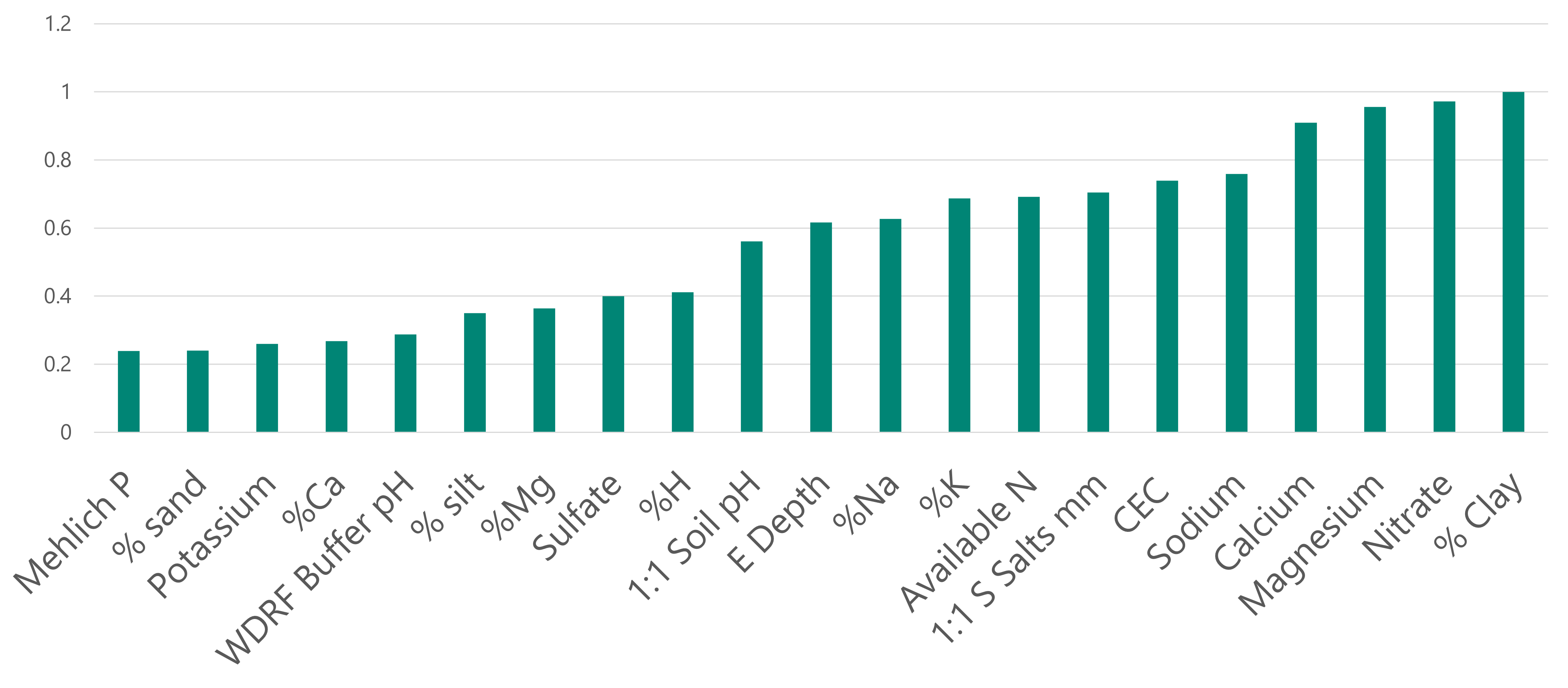}
    \caption{Standardized MSE Gain on Variable Removal}
    \label{fig:var-imp}
\end{figure}

\noindent \paragraph{Sensitivity Analysis}
In order to gain a better understanding of what variables are more impactful in improving generalization as auxiliary variables, we measure variable importance by variable removal. We leave one variable out at a time during training and evaluate how much test MSE increases due to removal. Figure~\ref{fig:var-imp} reports the standardized test MSE gain when each of the soil attribute variables is removed from the auxiliary training dataset. The higher the variable importance the more significant the influence on changes in OM. For instance, clay soils tend to have a higher capacity to retain moisture and nutrients, which in turn affects the decomposition of organic matter. Similarly, nitrate, a form of nitrogen readily available to plants, influences plant growth and increases OM through crop residues. This analysis helps shed light on which variables are important in improving out-of-sample generalization as auxiliary variables. The collection of data on these attributes may help farmers draw better insights about their farms. More results provided in the Appendix.


%% file: conclusion.tex
In this paper, we propose using a causal and contrastive constraint minimization mechanism to improve the estimation of organic matter (OM) from remote sensing data. This framework enables us to transfer knowledge from locations with soil information to locations where collecting this information may be infeasible, improving scalability to other regions in the world where data collection is cost-prohibitive. The sensitivity analysis may also help identify key soil characteristics influencing OM. Farmers can use the information on important soil characteristics to make targeted decisions on soil amendments. For example, if nitrogen levels are identified as crucial for organic matter, farmers might adjust their fertilizer application strategies. Moreover, if soil characteristics favoring organic matter are identified, farmers may consider reduced tillage practices to preserve soil structure and organic matter content. These efforts can help optimize soil management practices to allow more precision and efficient use of resources. There are several extensions to the study that can be explored. For example, although our study accounts for changes due to soil characteristics, it is also important to consider changes in OM due to weather dynamics. Two locations with similar soil texture can have widely varying weather and climate. Due to the limited availability of reliable soil attribute information, the use of more readily available weather data along with remote sensing data will be an important future extension of this work.

%% file: appendix.tex
\section*{Appendix}

\subsection*{Study Area}

The sites included in the study are obtained from the publicly available \href{https://www.genomes2fields.org/home/}{Genomes to Fields (G2F) dataset}. The locations of the sites are given in the Table~\ref{tab:site-names}. For these sites, the data set provides accurate and reliable soil attribute information. Modeling using these sites enables us to make accurate predictions. The G2F data contains information from 59 environments (where an environment refers to a location in a given year) over 6 locations. The pretrained encoder is pretrained using satellite imagery from all 59 environments on multiple days in the year. For the results on domain adaptation and generalization, the satellite imagery on the day of soil sampling is used as input to minimize the bias in input data since the organic matter values are available on the days when the soil was sampled in a given location in a year. Experiments on early predictions suggest that using satellite imagery data from January 1$^{\text{st}}$ of each year instead of the day of sampling also provide reasonable generalization. This enables to use the model even when we do not have information on when the soil was sampled in the year. 

\begin{table}[h]
    \centering
    \begin{tabular}{c|c|c|c|c|c}
    \toprule
         deh & gah&iah & ilh & mih & geh   \\
    \midrule
         Delaware & Georgia & Iowa & Illinois & Michigan & Germany \\
    \bottomrule
    \end{tabular}
    \caption{Site Names}
    \label{tab:site-names}
\end{table}

\begin{figure}[h] 
  \begin{center}
    \includegraphics[width=0.7\textwidth]{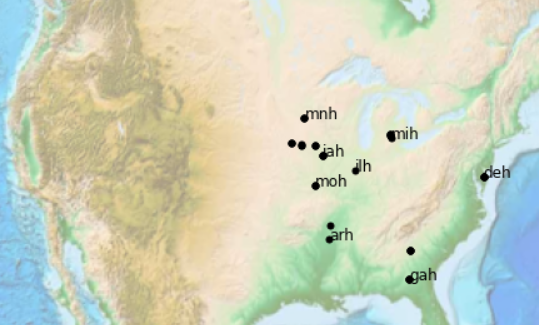}
  \end{center}
  \caption{Sites with Soil Attribute Information}
  \label{fig:map}
\end{figure}

\subsection*{Soil Atttributes Definition}

While the complete information on data collection, variable definition for all variables included in this study are given in \href{https://www.genomes2fields.org/home/}{Genomes to Fields (G2F) dataset}, we define the variables that are predicted to be important as auxiliary training variables.

\begin{table}[ht]
    \centering
    \scriptsize
    \begin{tabular}{c|c}
    \toprule
         Variable Name & Variable Definition  \\
     \midrule
        \% Clay & Percentage of clay composition in soil sample\\
         Nitrate & Available nitrate in parts per million (ppm)\\
         Magnesium & Available Magnesium in ppm\\
         Calcium & Available Calcium in ppn\\
         Sodium & Available sodium in ppm\\
         CEC& Cation Exchange Capacity (me/100g)\\
         1:1 S Salts mm& Soluble salts concentration in soil  \\
         Available N & Amount of nitrogen in pounds per acre \\
         \% K & Percentage of Potassium\\
         \% Na & Percentage of Sodium \\
         E Depth& Soil sample collection depth\\
         1:1 Soil pH & Soil pH level in a mixture, by weight, one-part soil to one-part distilled H2O \\
         \% H& Percentage of Hydrogen \\
         Sulfate& Available sulfate in ppm\\
         \% Mg& Percentage of Magnesium\\
         \% silt& Percentage of silt composition in soil sample\\
         WDRF Buffer pH & Woodruff method for measuring total soil acidity \\
         \% Ca & Percentage of Calcium \\
         Potassium & Available Potassium in ppm\\
     \bottomrule
         
    \end{tabular}
    \caption{Variable Definitions}
    \label{tab:vardef}
\end{table}

\subsection*{Differences in Environments}
\begin{figure}[h]
  \begin{center}
    \includegraphics[width=0.9\textwidth]{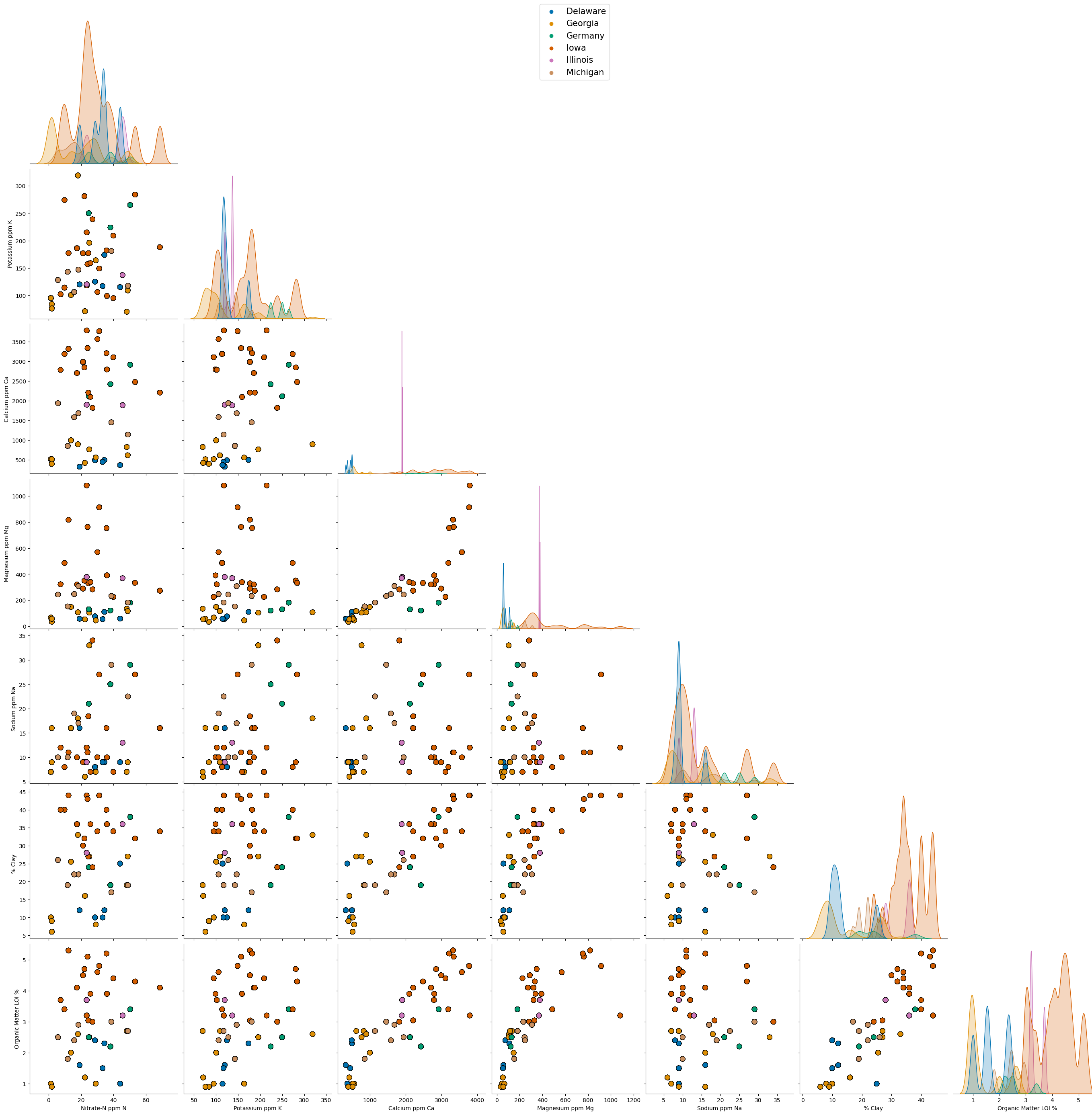}
  \end{center}
  \caption{Pairplot showing relationships between different soil variables.}
  \label{fig:pairplot}
\end{figure}

\begin{figure}[h]
  \begin{center}
    \includegraphics[width=0.3\textwidth]{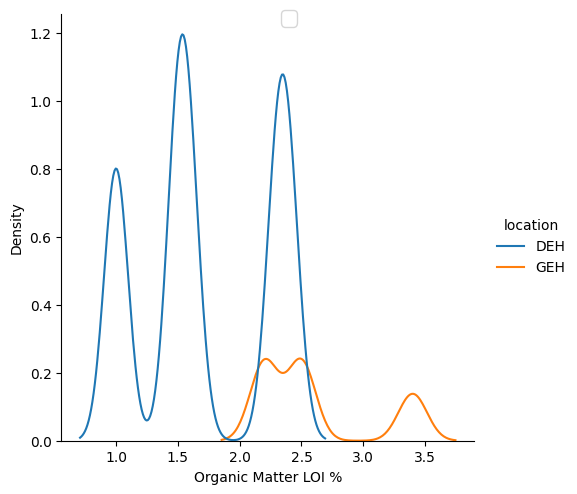}
  \end{center}
  \caption{Distribution of OM for Delware (DEH) and Germany (GEH) locations.}
  \label{fig:farthest}
\end{figure}

We also show the differences in the environments using pairplot in the Figure ~
 \ref{fig:pairplot}. Note in particular the difference in the OM distribution for Delaware and Germany in Fig. ~\ref{fig:farthest}. In Table~\ref{tab:finetune}, the results show that the proposed approach learns better even when fine tuned with a location farthest from the test location with distribution differences. 
 
\subsection*{Satellite Imagery processing}
We pre-process the raw Sentinel 2 imagery to remove clouds using the SpaceEye algorithm~\cite{spaceye}. The following bands are retained after pre-processing: B02, B03, B04, B05, B06, B07, B08, B8A, B11 and B12. For both the satellite imagery and soil attributes, we use min-max scaling to standardize the data.

\section*{Model Overview}

\subsection*{Backbone Model}
\begin{figure}[h]
    \centering
    \includegraphics[width=0.8\textwidth]{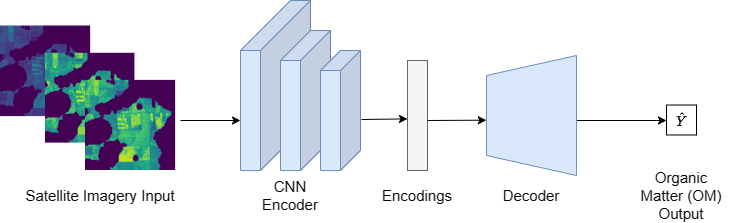}
    \caption{The backbone model is a CNN autoencoder. The input to the model is satellite imagery data and the response is organic matter value.The encoder includes 3 sets of convolution layers stacked with ReLU activation and max-pooling layers. The encoder maps the information to an array of encodings that are used as input to the decoder, which includes fully connected layers.}
    \label{fig:backbone-model}
\end{figure}

The backbone model is a CNN autoencoder showed in the Fig.~\ref{fig:backbone-model}.

\subsection*{Causal and Contrastive Constraint Minimization}


\subsection*{Causal Constraint Minimization}

Following the work constraints enforced in CACM \cite{kaur2022modeling}, the regularization constraints for attributes that are independent, caused by OM and confounded with OM, respectively, are as follows,

\begin{equation}
    \sum_{i=1}^{|A_{ind}|} \sum_{j>i}MMD(P(\phi(x)|a_{i,ind}),P(\phi(x)|a_{j,ind}))
\end{equation}

\begin{equation}
    \sum_{i=1}^{|A_{cause}|} \sum_{j>i}MMD(P(\phi(x)|a_{i,cause},y),P(\phi(x)|a_{j,cause},y))
\end{equation}

\begin{equation}
    \sum_{|E|}\sum_{i=1}^{|A_{conf}|} \sum_{j>i}MMD(P(\phi(x)|a_{i,conf}),P(\phi(x)|a_{j,conf}))
\end{equation}

\begin{figure}[h]
    \centering
    \includegraphics[width=0.9\textwidth]{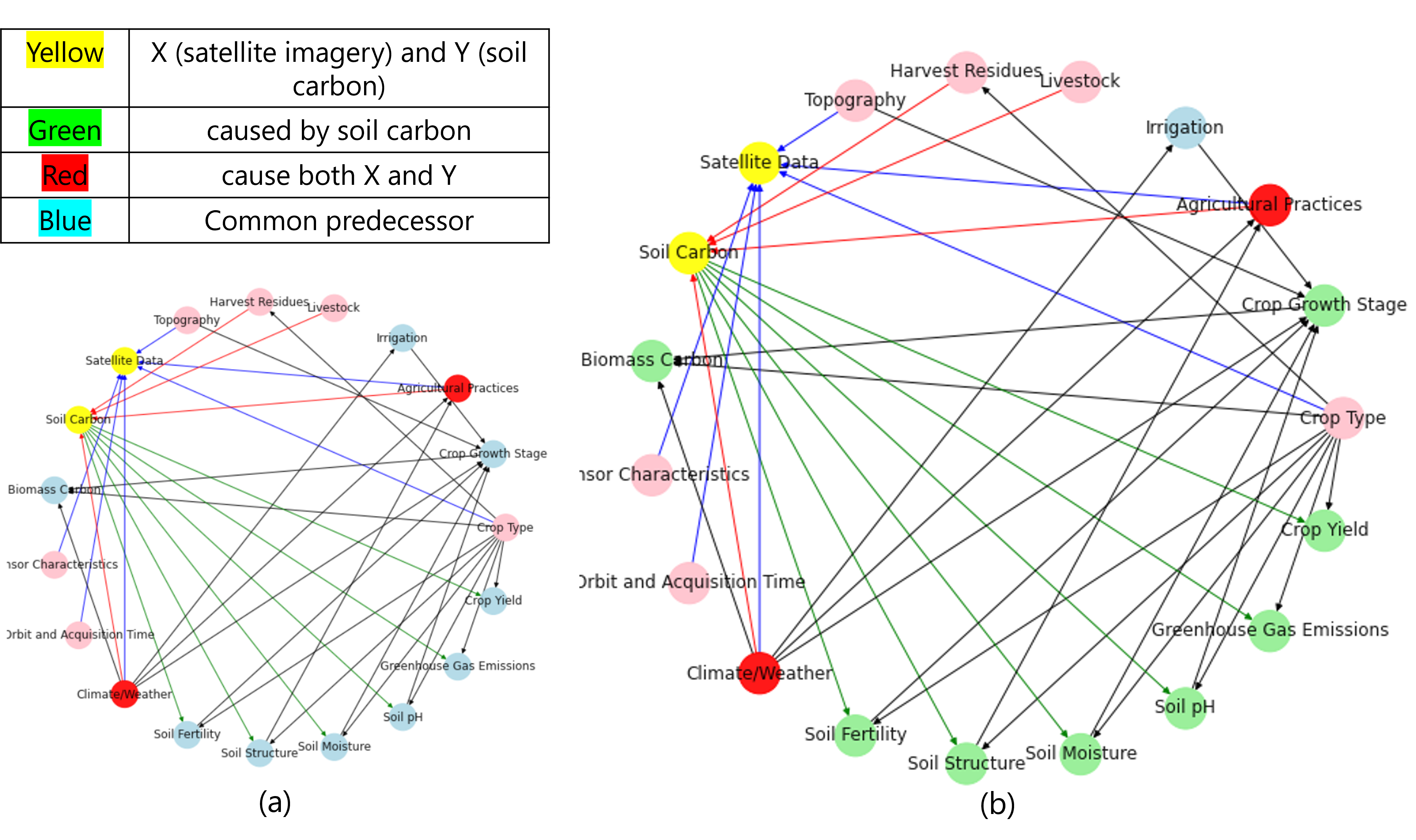}
    \caption{Causal graph among the soil attributes. Several attributes in the data are both caused by $Y$ and confounded with $Y$. (a) The sub-figure shows a version of the causal graph where these nodes are treated as confounded with $Y$. (b) The sub-figure shows a version of the causal graph where these nodes are treated as caused by $Y$. The empirical generalization performance for the graph in sub-figure (a) is better.  }
    \label{fig:causalgraph}
\end{figure}

More details on how to leverage CACM can be found in the work Kaur et al. \cite{kaur2022modeling}. In our work, we use the causal graph given in Figure~\ref{fig:causalgraph} to identify which constraints to enforce.

\subsection*{Results on Pre-training the Encoder}

\begin{figure}[h]
  \begin{center}
    \includegraphics[width=0.7\textwidth]{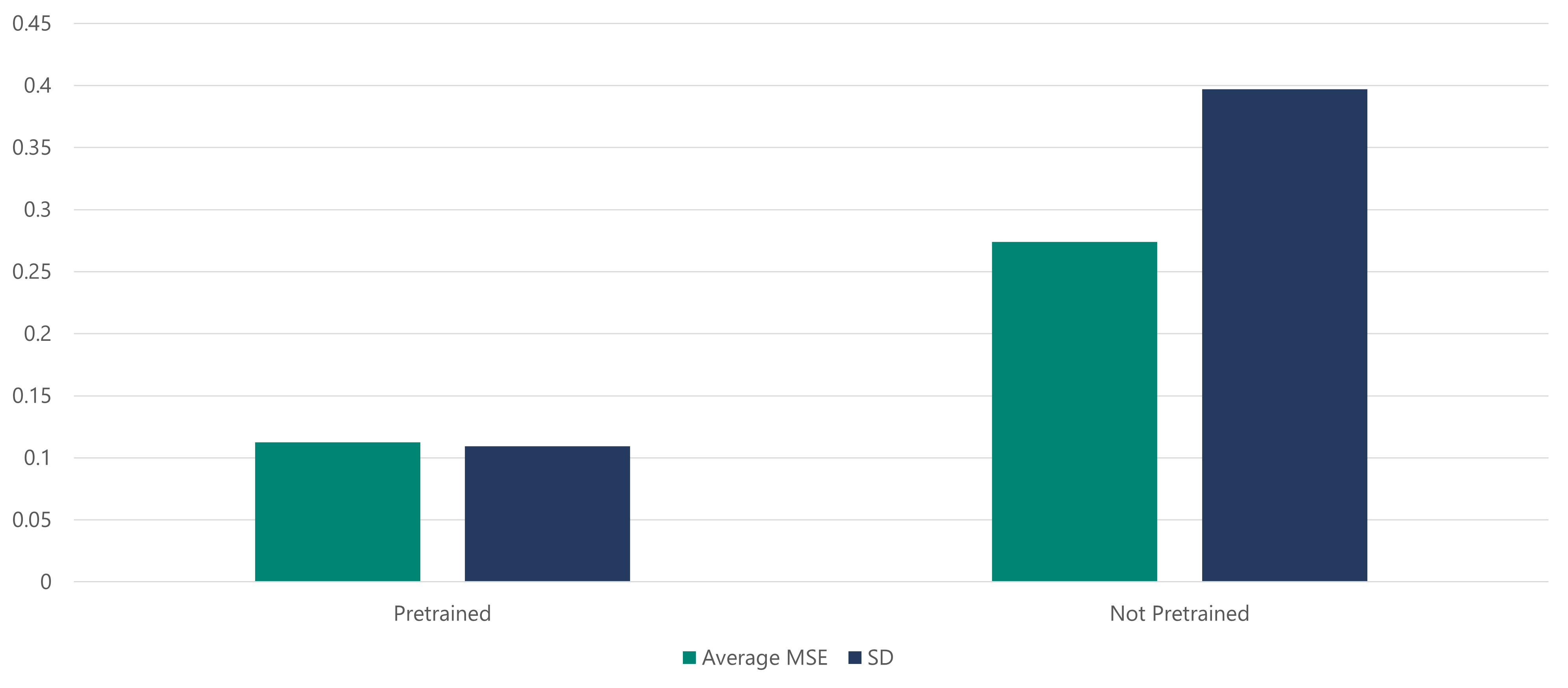}
  \end{center}
  \caption{Pretrained Encoder}
  \label{fig:pretrain}
\end{figure}

Figure~\ref{fig:pretrain} shows the change in test MSE when the encoder is pretrained using more satellite data from other time periods. Pretraining on a larger satellite dataset helps the encoder learn the underlying physical patterns that are present in the remote sensing data.


\subsection*{Results on CACM versus Encoding-based CACM}

In this study, we report the results in Fig. ~\ref{fig:cacmvsencoding} using encoding-based CACM, which regularizes the encoding space instead of the original variable space as was originally proposed. This reduces any over-smoothing in the output space. This further allows the decoder to focus on train parameters that only focus on OM estimation. We provide results on the domain adaptation experiments wherein we finetune the model on the farthest environment.

\begin{table}[h]
\centering
    \begin{tabular}{c|c}
    \toprule
    Model & MSE \\
    \midrule
         CNN$_{\text{CACM}}$& 0.1134  \\
         CNN$_{\text{Original CACM}}$& 0.1308 \\ 
    \bottomrule 
    \end{tabular}
    \caption{MSE comparison between original CACM and encoding based CACM.}
        \label{fig:cacmvsencoding}

\end{table}

\subsection*{Results on Sensitivity Analysis}

Table~\ref{tab:var-rank} also provides a comparison of variable ranking obtained from this sensitivity analysis with ranking for similar analysis when the soil attributes are used as input data.

These rankings are in agreement for several of the variables, such as percentage of clay (\% Clay) in soil, magnesium (ppm) in soil, and CEC. However, variables that have been known to significantly impact organic matter content, such as nitrate, have a disagreement in ranking when used as auxiliary variables and as input variables. This agreement may arise since the ranking derived based on auxiliary variables also accounts for input data satellite imagery that takes care of other unobserved variables whose impact we are unable to measure otherwise. The disagreement may also arise because of a difference in the role of the variables during training - while the impact of the input variable is directly mapped to response, the auxiliary variable in our framework is used indirectly to influence the training process. Interestingly, some variables, like the percentage of potassium in the soil, which are important in determining changes in organic matter, do not come out as important as the input variable. This may be due to the effect of unobserved confounders or noise in data since the correlation of this variable with OM is also relatively lower.


\begin{table}[h]
    \centering
    \begin{tabular}{c| c| c}
    \toprule
        Variable &  Importance as auxiliary variable  & Importance as input variable   \\
        \midrule
         \% Clay & 1 & 1\\
         Nitrate & 2 & 14 \\
         Magnesium & 3 & 3 \\
         Calcium & 4 & 10 \\
         CEC & 6 & 4\\
         \% Potassium & 9 & 27 \\
         Available Nitrogen & 8 & 13 \\
         \bottomrule
    \end{tabular}
    \caption{Variable Ranking by Descending Importance}
    \label{tab:var-rank}
\end{table}
